\documentclass[sigconf]{acmart}
\acmSubmissionID{589}

\usepackage{bm,balance}

\newcommand{\fig}[1]{Figure~\ref{#1}}

\newcommand{\norm}[1]{\left\lVert#1\right\rVert}

\long\def\ignorethis#1{}

\makeatletter
\DeclareRobustCommand\onedot{\futurelet\@let@token\@onedot}
\def\@onedot{\ifx\@let@token.\else.\null\fi\xspace}

\makeatother
\def\shortcite#1{ [\citeyear{#1}]}

\renewcommand{\eqref}[1]{Equation~(\ref{eq:#1})}

\newcommand{\argmin}{\operatornamewithlimits{argmin}}

\copyrightyear{2022} 
\acmYear{2022} 
\setcopyright{acmlicensed}
\acmConference[SA '22 Conference Papers]{SIGGRAPH Asia 2022 Conference Papers}{December 6--9, 2022}{Daegu, Republic of Korea}
\acmBooktitle{SIGGRAPH Asia 2022 Conference Papers (SA '22 Conference Papers), December 6--9, 2022, Daegu, Republic of Korea}
\acmPrice{15.00}
\acmDOI{10.1145/3550469.3555428}
\acmISBN{978-1-4503-9470-3/22/12}

\begin{document}
\title[Transformer Inertial Poser]{Transformer Inertial Poser: Real-time Human Motion Reconstruction from Sparse IMUs with Simultaneous Terrain Generation}

\author{Yifeng Jiang}
\affiliation{%
  \institution{Stanford University}
  \country{United States of America}
}
\email{yifengj@stanford.edu}

\author{Yuting Ye}
\affiliation{%
  \institution{Meta Reality Labs Research}
  \country{United States of America}
}
\email{yuting.ye@fb.com}

\author{Deepak Gopinath}
\affiliation{%
  \institution{Meta Reality Labs Research}
  \country{United States of America}
}
\email{dgopinath@fb.com}

\author{Jungdam Won}
\affiliation{%
  \institution{Meta Reality Labs Research}
  \country{United States of America}
}
\email{jungdam@fb.com}

\author{Alexander W. Winkler}
\affiliation{%
  \institution{Meta Reality Labs Research}
  \country{United States of America}
}
\email{winklera@fb.com}

\author{C. Karen Liu}
\affiliation{%
  \institution{Stanford University}
  \country{United States of America}
}
\email{karenliu@cs.stanford.edu}



\begin{abstract}

Real-time human motion reconstruction from a sparse set of (e.g. six) wearable IMUs provides a non-intrusive and economic approach to motion capture. Without the ability to acquire position information directly from IMUs, recent works took data-driven approaches that utilize large human motion datasets to tackle this under-determined problem. Still, challenges remain such as temporal consistency, drifting of global and joint motions, and diverse coverage of motion types on various terrains. We propose a novel method to simultaneously estimate full-body motion and generate plausible visited terrain from only six IMU sensors in real-time. Our method incorporates 1. a conditional Transformer decoder model giving consistent predictions by explicitly reasoning prediction history, 2. a simple yet general learning target named "stationary body points” (SBPs) which can be stably predicted by the Transformer model and utilized by analytical routines to correct joint and global drifting, and 3. an algorithm to generate regularized terrain height maps from noisy SBP predictions which can in turn correct noisy global motion estimation. We evaluate our framework extensively on synthesized and real IMU data, and with real-time live demos, and show superior performance over strong baseline methods.

\end{abstract}

%
%


%
%



\maketitle

\section{Introduction}

Real-time reconstruction of 3D human motion is crucial for applications in various domains, such as biomechanics and sports analysis, motion-based video games, and virtual presence in VR/AR systems. While marker-based optical motion capture systems~\cite{vicon} remain an ideal option for research labs and professional studios due to the superior accuracy, more and more applications demand a portable, less costly, and minimally-intrusive mocap system that reconstructs human movements in real-time and can be used anywhere by everyone.

\begin{figure}[h]
  \includegraphics[width=1.0\linewidth]{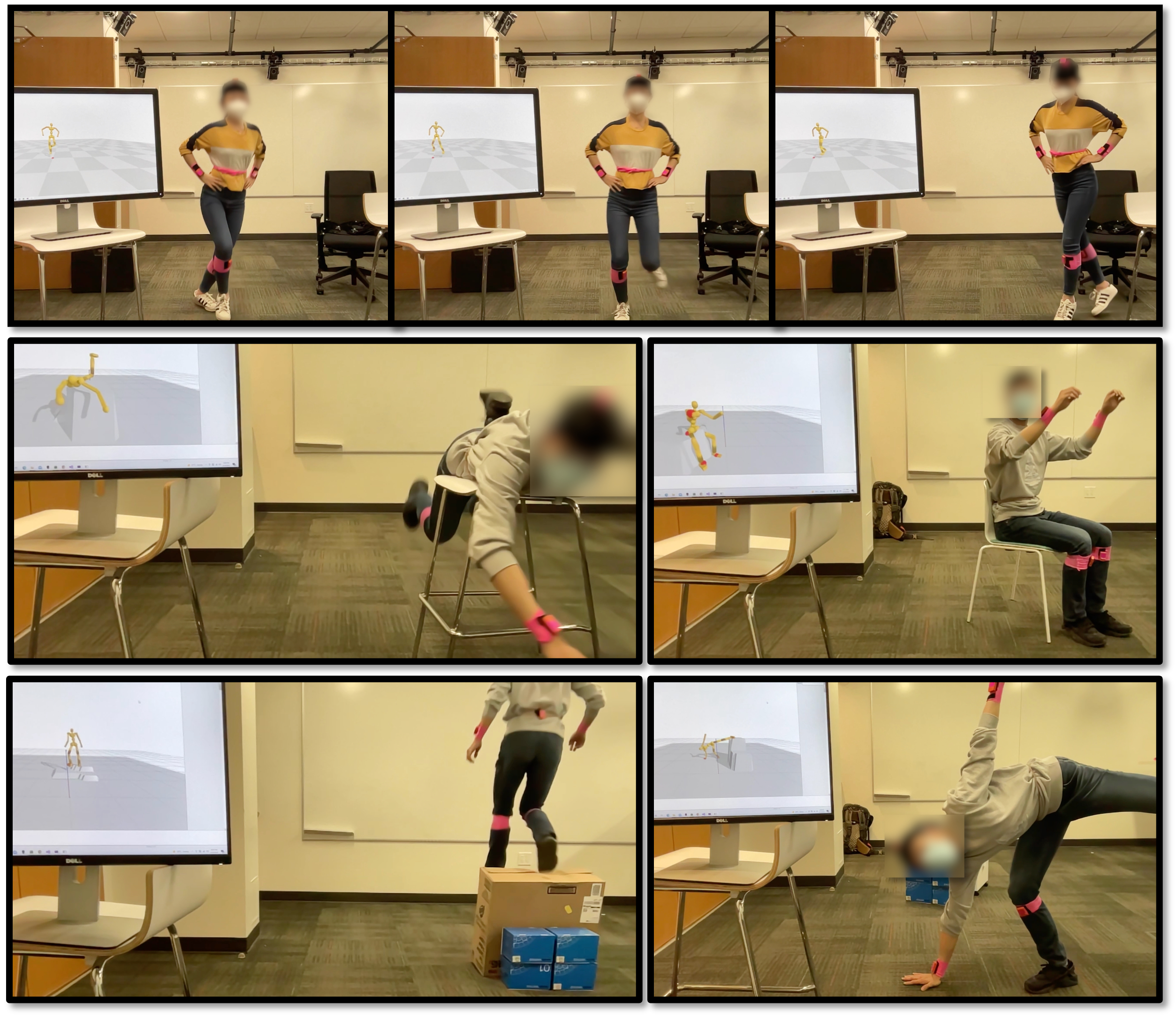}
  \caption{We develop an attention-based deep learning method to reconstruct full-body motion from six IMU sensors in real-time. Our method simultaneously generates plausible terrain maps that can explain the reconstructed motions, for a variety of motion types.}
  \label{fig:swinging_max}
\end{figure}

Among many sensing modalities, such as RGB cameras \cite{Kanazawa:2019:CVPR,OpenPose:PAMI:2019,Densepose:CVPR:2018}, depth cameras \cite{Taylor:CVPR:2012,Wei:TOG:2012}, or wearable electromagnetic sensors \cite{Kaufmann:ICCV:2021}, inertial measurement sensors (IMUs) \cite{Marcard:SIP:2017,DIP:SIGA:2018} provide many unique advantages. IMU-based mocap is self-contained, egocentric, and untethered, applicable to both indoor and outdoor activities in all lighting conditions. For example, an IMU system can capture long-distance motion of a hiker or a mountain climber, but third-person cameras would be limited by their field of view. In addition, IMU-based mocap is unsusceptible to occlusion and is less sensitive to privacy issues. Recent works \cite{TransPose:SIG:2021, DIP:SIGA:2018} have advanced the field from commercial systems using 17 IMUs \cite{xsens} to machine-learning-based systems using only six IMUs, demonstrating the great potential of the IMU-based approach as a practical and reliable solution to "everyday motion capture".



With all the recent progress, there are still a few critical issues, such as temporal inconsistency of prediction, and global translational drifts, that prevent IMU-based mocap systems to achieve their full potential. In response to these challenges, we propose a novel method to enhance the reconstruction of full-body human motion in real-time from only six IMU sensors. First, inspired by recent success of Transformer models for free-form natural language generation \cite{radford:2018:improving}, we cast the IMU reconstruction as a constrained motion generation problem and solve it by learning a conditioned Transformer decoder model. Comparing with RNN-based models used by previous methods, our model is able to achieve more accurate joint angle estimation, especially for those motions with almost identical IMU signals such as sitting and standing, thanks to Transformer's improved ability to reason about its own past predictions by taken them explicitly as input. 

Second, we introduce a general technique to address the gradual drift of global translation and joint motion in a unified way. We train our Transformer decoder model to additionally predict the points on the character that have near-zero velocity, which we call "Stationary Body Points" (SBPs). Once learned, SBPs can be used by analytical routines to counteract drifts in motion, which is particularly helpful when there is a distribution mismatch between test and training sets due to noisy or corrupted IMU sensors, or unseen motions. Concurrent to our work, \cite{PIP:CVPR:2022} also tackles the drifting problems using a combination of physics-based optimization and improved initialization techniques for RNNs. Our work is interestingly complementary to \cite{PIP:CVPR:2022} on this front. 

The IMU-based approach opens the door to motion capture in large environments with a variety of terrains, posing a new research challenge of terrain reconstruction in real-time. As our aforementioned motion estimators are terrain-agnostic, by leveraging the predicted SBPs, our method can be used to generate plausible height maps of the terrain traversed by the human. Moreover, instead of treating terrain reconstruction as a byproduct of our method, we simultaneously estimate the human motion and the corresponding terrains such that the generated terrains can also be used to regularize the reconstructed global motion in real-time.


We evaluate our system, which we call "Transformer Inertial Poser" (TIP), extensively on synthetic and real datasets, and showcase clear improvement over recent state-of-the-art methods on flat terrains. As no real IMU datasets with non-flat terrains exist for training or testing, we qualitatively evaluate our reconstructed motions along with the simultaneously generated terrains both in simulation and with our own live demos. We additionally show real-time live demos covering a wide variety of motion types, many of which unseen in previous works.

\section{Related Work}

Human motion reconstruction from various sensor inputs has been studied for a long time particularly in Computer Graphics and Computer Vision communities.  We mainly review prior works that use IMU sensors as part of or the sole input modality.  We also review motion generation models based on the Transformer~\cite{Transfomer:NIPS:2017} as it constitutes the core of our reconstruction model.  

A typical IMU sensor includes an accelerometer measuring 3-axis linear acceleration, a gyroscope measuring 3-axis angular velocity, and a magnetometer identifying the vector pointing Earth’s magnetic north. From these raw signals, sensor fusion algorithms based on Kalman filter or its extended version are used to provide more robust measures of the orientation~\cite{Vitali:JSEN:2021,Rosario:JSEN:2018,Bachmann:VRST:2001,Foxlin:VRAIS:1996}.  IMU sensors have been used along with vision-based sensors such as RGB or RGB-D cameras for motion estimation. Some work regard IMU signals as extra constraints to regularize motions predicted from vision, where those constraints are formulated either in offline optimizations~\cite{Marcard:PAMI:2016,Marcard:ECCV:2018,Pons-Moll:ICCV:2011,Pons-Moll:CVPR:2010,Helten:ICCV:2013,Zheng:ECCV:2018}, online (often per-frame) optimizations~\cite{Malleson:IJCV:2020,Malleson:3DV:2017,Zhang:CVPR:2020}, or learning deep neural networks~\cite{Gilbert:IJCV:2019,Trumble:BMVC:2017}.  There have also been works combining IMUs with other modalities such as optical markers~\cite{Andrews:CVMP:2016} or ultrasonic~\cite{Liu:I3D:2011,Vlasic:SIG:2007}. Although these systems can produce plausible motions, they also suffer from inherent limitations of vision-based sensors such as narrow capture region, occlusion, or sensitivity to the light condition.

With IMU sensors getting more compact and inexpensive, they have received increasing attentions from both industry and research communities for a standalone body tracking solution. Popular commercial products such as \cite{xsens} and \cite{rokoko} can generate high-quality human motions ready to be used in real-time game engines. However, requiring a sophisticated full-body setup with at least 17 IMUs hinders their accessibility to everyday users. Researchers have therefore proposed body tracking systems with a small number of IMUs sparsely placed on the body, usually utilizing statistical body models and/or high quality optical mocap data as prior to mitigate input signals being under-specified. Marcard et al.~[\citeyear{Marcard:SIP:2017}] developed an offline system (SIP) with only six IMUs, which optimizes poses and the parameters of the SMPL body model~\cite{SMPL:SIGA:2015} to fit the sparse sensor input. Huang et al.~[\citeyear{DIP:SIGA:2018}] learned a deep neural-net model (DIP) from a large amount of motion capture data to directly map the IMU signals to poses. Their model is based on bidirectional recurrent neural networks (BRNN), so the system can run in an online manner while considering both the past and future sensor inputs with a negligible latency, outperforming previous non-learning online methods. An ensemble of BRNNs was further adopted by Nagaraj et al.~\shortcite{Nagaraj:APPIS:2020} to improve upon the results. However, the two real-time solutions mostly focus on reconstructing the local joint motion without global translation. Yi et al.~[\citeyear{TransPose:SIG:2021}] proposed a new neural model (TransPose) where the progressive upscaling of joint position estimation showed more accurate pose estimation. The model can additionally generate accurate global root motions by combining a supporting-foot heuristics and a small learned deep network, similar to \cite{rempe2021humor}. Recently, an extension of this system (PIP)~[\citeyear{PIP:CVPR:2022}] has been introduced, which is concurrent to our paper, where the predicted motions are further optimized to reduce violations of physics laws \cite{PhysCap:SIGA:2020}. We explore this problem domain with a set of drastically different techniques, and with much more relaxed assumptions on the environment geometry, while producing comparable or better reconstruction results.

The blooming AR/VR industry draws attention to full-body motion synthesis, using IMUs on headsets and controllers. With no senors available on the lower body, utilizing deep reinforcement learning, Luo et al.\shortcite{DRK:NIPS:2021} is able to synthesize physically valid locomotion from only the 6D egocentric headset pose. Cha et al.~\shortcite{MEH:VRUI:2021} complements pose estimation from headset cameras with IMUs when the hands are out-of-view. Choutas et al.\shortcite{Choutas:arxiv:2021} and Dittadi et al.\shortcite{dittadi:ICCV:2021} experimented with deep generative models conditioned on headset and controllers poses to synthesize full-body poses. More similar to our work is LoBSTR~\cite{Yang:CGF:2021}, where they include an IMU on the waist in addition to IMUs on the headset and controllers. Using a recurrent network, they can synthesize both sitting and running motions from only 4 sensors. We opt to use 6 IMUs with lower body information for accurate motion reconstruction rather than synthesis, but these sparser setups are fruitful future directions.




Since the inception of the Transformer, attention-based models have become the state-of-the-art on many problems involving sequence data, such as language translation~\cite{GPT-3:arXiv:2020} and audio generation~\cite{Jukebox:arXiv:2020}.  It is natural to also apply Transformer models to synthesizing human motion. Aksan et al.~\shortcite{Aksan:3DV:2021} developed a generative model using dual attention mechanism to capture spatial and temporal correlations, which predicts future full-body locomotion given a short history.  Petrovich et al.~\shortcite{Petrovich:ICCV:2021} used a Transformer and a variational autoenoder conditioned on action labels, such as  \textit{walking} or \textit{jumping}, to generate full body motions. Valle-P\'{e}rez et al.~\shortcite{Valle-Perez:SIG:2021} instead combined a Transformer with normalizing flows to synthesize dancing motion from music, building on a similar prior work~\cite{Li:ICCV:2021}. For motion reconstruction, Kim et al.~\shortcite{Kim:PRL:2021} experimented with a Transformer encoder-decoder model with sparse synthetic input features, and found it more effective than recurrent networks. In our case, we face the additional challenge of handling noise from real IMU sensors.

\section{Transformer Inertial Poser (TIP)}

\begin{figure}[ht]
\centering
\includegraphics[width=\linewidth]{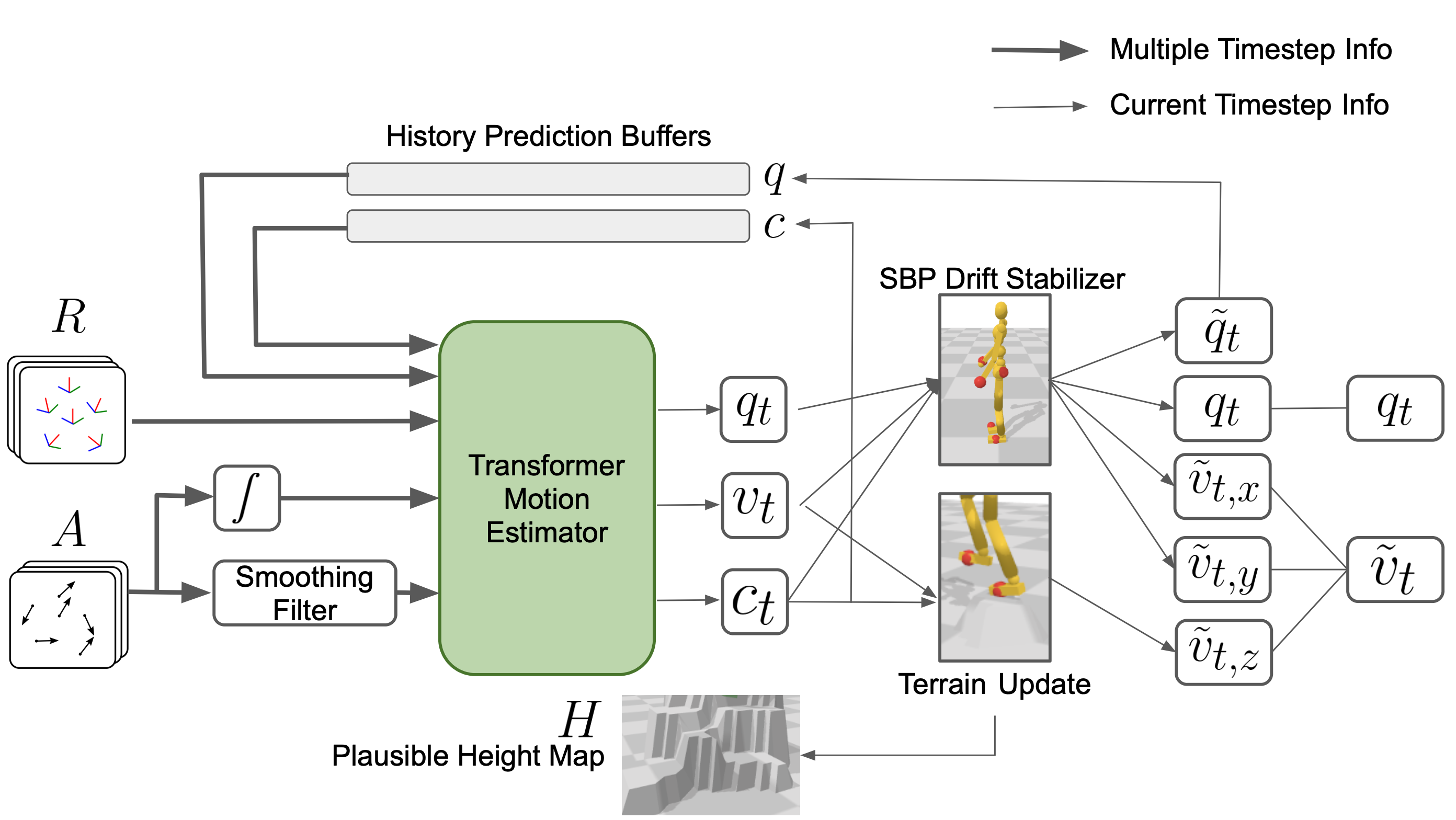}
\caption{Overview of our pose estimation algorithm. Thicker arrows are "pipes" flowing multi-timestep information at each model step, while thinner arrows flow current-step information.}
\label{fig:overview}
\end{figure}

We introduce a real-time human motion reconstruction technique from six IMU sensors placed on the user's legs, wrists, head, and waist. Our approach combines a learning-based model and an analytical routine to estimate full-body joint angles $\bm{q}$ and the root velocity $\bm{v}$ from a real-time stream of IMU orientation $\bm{R}$ and acceleration $\bm{A}$ signals. Summarized in \fig{fig:overview}, our method depends on a learned \emph{Transformer decoder} to estimate the motion and an analytical \emph{drift stabilizer} to refine the estimation. In addition to $\bm{q}$ and $\bm{v}$, the Transformer decoder also predicts stationary body points (SBPs) $\bm{c}$ which are used by the drift stabilizer during run time to improve the accuracy of the reconstructed motion. The drift stabilizer utilizes the predicted SBPs to mitigate drifts in motion over time and a non-learning \emph{terrain update} module estimates the corresponding terrain while further improving motion reconstruction. A plausible terrain height map is generated and updated in real time, as an additional product of our algorithm.

\begin{figure}[ht]
\centering
\includegraphics[width=\linewidth]{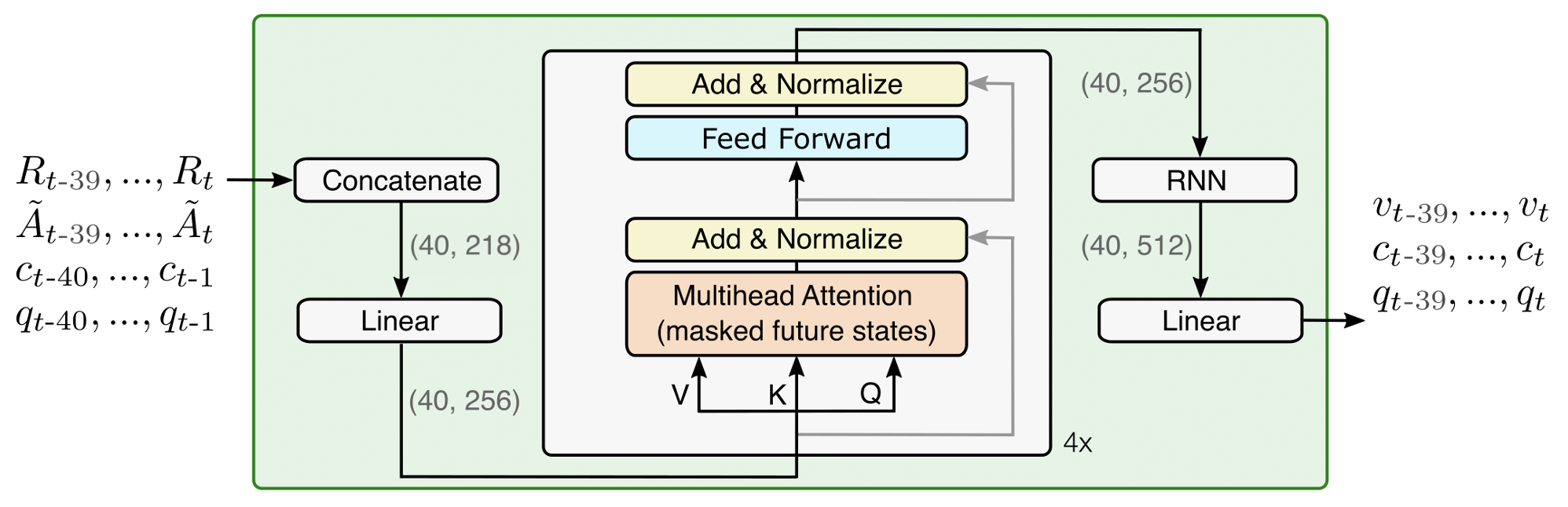} 
\caption{Our conditional Transformer decoder architecture during training, showing $M=39$. Following GPT training conventions, during training, input and output contain the same (but time-shifted by 1) ground-truth motion ($\bm{q}~\&~\bm{c}$). During testing, output from $t-39$ to $t-1$ are discarded and the last ones $\bm{v}_t, \bm{c}_t, \bm{q}_t$ are used as current prediction, and input ($\bm{q}~\&~\bm{c}$, i.e. the history buffers in \fig{fig:overview}) is autoregressively updated by most recent outputs $\bm{v}_t, \bm{c}_t$ at each step.}
\label{fig:transformer}
\end{figure}

\subsection{Transformer Motion Estimator}
\label{sec:method-transformer}

To simplify the notations, consider the problem of reconstructing only the joint angles $\bm{q}$. The standard learning-based approach can be summarized as $\max P(\bm{q}_t | \bm{R}_{t-N:t}, \bm{A}_{t-N:t})$, where a neural network is trained to output the most probable current joint angle given the IMU readings. Note that since 6 noisy IMUs do not provide enough signals to fully determine the whole-body motion, we follow previous work \cite{DIP:SIGA:2018} to include a time window from $t-N$ to $t$, mitigating under-specification. 

Such models could still have trouble distinguishing motions with similar IMU readings over the entire window, such as sitting vs. standing, and the transition in between (if slow). To principally address the issue, we draw an analogy between motion generation and language modeling: $\max P(\bm{q}_{t-M:t})$, where the goal is to free-form generate all motion (language) sequences that are natural. Applying recursively, a neural network can be used to generate plausible $\bm{q}_t$ conditioned on all the previous generations from itself:
\begin{equation*}
    \max P(\bm{q}_{t-M:t}) = \max P(\bm{q}_{t} | \bm{q}_{t-M:t-1}) P(\bm{q}_{t-1} | \bm{q}_{t-M:t-2}) \cdots P(\bm{q}_{t-M}).
\end{equation*}
A Transformer decoder model \cite{Transfomer:NIPS:2017, radford:2018:improving} excels at this task as it explicitly takes previous predictions as input and is capable of reasoning about such contexts fast enough in parallel with temporal attention mechanisms. The Transformer decoder model is able to take variable-sized history as input but in practice, a maximum window length $M$ is usually set.

Unlike free-form language generation \cite{radford:2018:improving}, we have additional constraints from the IMU sensors. As such, we model our constrained motion generation problem with a \textit{conditional} Transformer decoder, i.e. $\max P(\bm{q}_{t-M:t} | \bm{R}_{t-M:t}, \bm{A}_{t-M:t})$, where we feed the model the sequences of IMU readings $\bm{R}$ and $\bm{A}$ in parallel to its past predictions, with the same sequence length at each step, up to a maximum of $M$.

\paragraph{Model.} The input to our model includes the IMU acceleration readings after smoothing filtering or integration $\tilde{\bm{A}} \in \mathbb{R}^{36 \times (M+1)}$ (Appendix A), and the IMU orientations $\bm{R} \in \mathbb{R}^{54 \times (M+1)}$ represented as flattened rotation matrix (length 9) from 6 sensors (\fig{fig:transformer}). The output includes 18 joint angles $\bm{q}_t \in \mathbb{R}^{108}$ defined in the SMPL \cite{SMPL:SIGA:2015} human model (excluding joints such as toes following previous works), the root linear velocity, $\bm{v}_t \in \mathbb{R}^{3}$, and the stationary body points (SBPs) $\bm{c}_t \in \mathbb{R}^{20}$. Each joint in $\bm{q}_t$ is represented redundantly as first two columns of its local rotation matrix for unique and numerically stable ground-truth labels \cite{zhou2019continuity}. The root orientation is given directly by one of the IMUs placed on the waist. We empirically found that adding an recurrent layer to "summarize" output embedding of the Transformer accelerates convergence during training.

Following the standard practice of Transformers (GPT \cite{radford:2018:improving}) training, given the shifted ground-truth sequences from $t-M-1$ to $t-1$, the model learns to predict \textit{in parallel} the whole sequence from $t-M$ to $t$ for efficiency (\fig{fig:transformer}). To prevent the model from learning simply to shift the input by one timestep, a causal mask \cite{Transfomer:NIPS:2017} is added to hide future attention information, mimicking the test-time setting. During test time, since we only care about the most recent prediction, output from $t-M$ to $t-1$ will be discarded at each prediction step.

There is still a large asymmetry between training and testing times. During training, the model sees ground-truth motion as the history input, but during testing, the history is noisily accumulated from its own predictions. Previous works, such as \cite{radford:2018:improving}, found in practice that such "teacher-forcing" training will not cause overfitting to the clean history. In our case, different from language where neighboring words are distinct, neighboring poses are usually similar to each other, providing much duplicate information more susceptible to overfitting. We add a 80\% dropout \cite{Srivastava:2014:DSW} to the history of $\bm{c}$ and $\bm{q}$, effectively dropping 4 frames out of 5. We also found that excluding $\bm{v}$ from history is important to prevent test-time autoregressive divergence, possibly because $\bm{v}$ is usually nearly constant in a time window, and the model could easily exploit and overfit to history $\bm{v}$ without truly reasoning about IMUs or history joint angles.

\subsection{Stabilizing Drift with SBPs}
\label{sec:method-sbp}
Combating drifts is one of the biggest challenges for IMU-based motion reconstruction. Unlike motion estimation methods using external cameras, IMUs have no direct sense of relative position, and learning or optimization based algorithms can all to some extent be seen as relying on double integration of the noisy acceleration readings to estimate position. If there is any biased error due to calibration or environment interference, the drifts of root translation or joint angles over time will result in artifacts, such as frequent foot skating or locking, or erroneous motion transitions.

\begin{figure}[h]
\centering
\includegraphics[width=\linewidth]{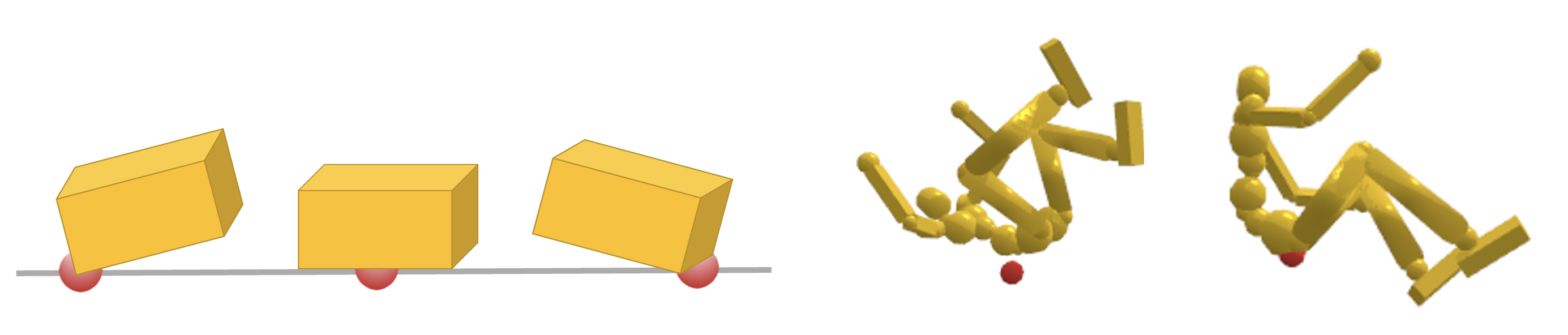}
\caption{During heel-to-toe contact in locomotion, contact patch is stationary but the foot link velocity is not zero. Similarly, a person rolling on the ground can be seen as rotating around changing stationary points.}
\label{fig:method-sbp}
\end{figure}

We present a simple and general technique, combining learned and analytical components, which does not require body-part-specific heuristics, nor rely on assumptions of the environment terrains. Besides motion predicting, our model will also predict Stationary Body Points (SBPs) $\bm{c}_t$ on the character, the representative locations on the human body with near-zero velocity. For example, the heel-to-toe rolling contact during walking results in moving SBPs across the foot, and rolling on the floor motion results in SBPs moving across the lower back and pelvis (\fig{fig:method-sbp}). Once learned, SBPs could be used for analytical correction of the predicted motion. Though the learning of SBPs could also be subject to inaccuracies, the analytical usage of the SBPs constrains the amount of errors they can produce.

\paragraph{Discover Ground-truth SBPs for Learning.} SBPs can be discovered on any body part, but in practice we choose the hands, feet, and pelvis to be the only candidate regions for SBPs in our implementation. We parameterize each SBP location as a vector offset $\bm{r}_t \in \mathbb{R}^3$ from the center of mass of the body part it belongs to. We use one additional bit for each SBP $b_t \in \{0,1\}$ to indicate the existence of SBP at the current moment. Therefore, SBP is represented as $\bm{c} \in \mathbb{R}^{20}$, where $\bm{c_t} = [b_t^{(i)}, \bm{r}_t^{(i)}]_{i=1,\cdots, 5}$.

We devise an efficient sampling-based method to search for the point on the rigid body with minimal velocity,
\begin{equation*}
\argmin_{\bm{r}} \norm{\bm{\omega} \times \bm{R}_B \bm{r} + \bm{v}},    
\end{equation*}
where $\bm{\omega}$ and $\bm{v}$ are angular and linear velocities of the body part, and $\bm{R}_B$ is the orientation of the body, which can all be easily obtained from ground-truth $\bm{q}_t, \bm{v}_t$ in training data. If the minimal velocity is below a small threshold, a SBP for that body part is found. 

We evaluate the velocity of every points in a 3D grid that encloses the body part of interest and find the point with the minimal velocity. For example, the 3D grid for a foot is a 6cm-thick box covering the bottom of the foot and that for the pelvis covers a larger region on the back. Evaluation for all candidate points can be done in parallel efficiently using matrix operations.

In practice, directly searching for the point with minimal velocity can lead to jittery SBPs over time, due to noise in motion capture data \cite{le2006robust} and approximating human body parts with rigid bodies. We thus add a temporal regularizer to the search criteria:

\begin{equation*}
    l(\bm{r}) = \norm{\bm{\omega} \times \bm{R}_B \bm{r} + \bm{v}} + 0.3 \norm{\bm{r} - \bm{r}_{t-1}},
\end{equation*}
where $\bm{r}_{t-1}$ is the solution in the previous frame, if existing. If minimal $l(\bm{r}^*)$ is less than a manually chosen threshold ($0.25$), we label $\bm{c}_t(k) = [1, \bm{r}^*]$. Otherwise, $\bm{c}_t(k) = [0, (0,0,0)]$. 

We did not find our results sensitive to the choices of regularization weight ($0.3$) or threshold, and use the same formula for all five SBPs, across the entire training dataset with various terrains.


\paragraph{Run-time Root Correction.} The predicted SBPs provide constraints to correct the root velocity. We set the corrected root velocity to be the average of $\tilde{\bm{v}}^{(i)}_t$, where each $\tilde{\bm{v}}^{(i)}_t$ is the root velocity that makes $i$-th (active) SBP exactly stationary in world space at current step. In practice, we only use this technique to correct root translation in the horizontal plane, and leave root correction in the height direction to the terrain estimation module described in Sec. \ref{sec:method-terrain}.

\paragraph{Run-time Joint Correction.} We can also modify the joint angle estimation $\bm{q}_t$ using SBPs. When a pair of SBPs are on for consecutive frames, we can solve for the joint angle correction that maintains the distance vector between the SBPs. We adopt two-bone inverse kinematics (IK) \cite{LMM-Holden}, which gives numerically stable solutions and changes only a minimal number of joints, $\tilde{\bm{q}}_t$. Directly replacing the joint angles $\bm{q}_t$ estimated by the Transformer decoder with $\tilde{\bm{q}}_t$ can result in discontinuity in motion when SBPs switch on and off. The problem is more visible as SBP onsets are imperfect model predictions. We adopt a new "soft-IK" technique utilizing the Transformer decoder's dependency on its own history of prediction. Intuitively, next model predictions will be improved if the history buffer is filled with more accurate motion. As such, our idea is to accept the current prediction $\bm{q}_t$ from the Transformer, but feed the corrected joint angles, $\tilde{\bm{q}}_t$, back to the Transformer's history buffer. This essentially creates a soft-IK constraint which does not enforce SBP pairs immediately in the current time step, but do so gradually in subsequent frames. 



\subsection{Plausible Terrain Generation}
\label{sec:method-terrain}

Inspired by SLAM (Simultaneous Localization And Mapping) algorithms~\cite{slam:whyte:2006,HPS:CVPR:2021}, our method generates a plausible terrain consistent with the reconstructed motion. In our case, simultaneously predicting the terrain and the motion is mutually beneficial in achieving improved results for both. The algorithm takes as input current pelvis and feet SBP locations, as well as Transformer-estimated vertical root velocity $\bm{v}_{t,z}$, and outputs a height map $\bm{H} \in \mathbb{R}^{L \times L}$ and the proposed vertical root correction $p \in \mathbb{R}$ ($\tilde{\bm{v}}_{t,z} = \bm{v}_{t,z} + p$). We do not consider hand SBPs here because in most cases it is hard to know if hand rests on something or simply stays stationary in the air.

This algorithm is based on two mild assumptions, albeit crucial to regularizing noisy SBPs from real IMU data. First, if two SBPs are horizontally nearby ($<1$m) and have sufficiently similar heights ($<0.1$m), we assume they have the same height on the terrain. As such, we assume there is no gradual slope in the scene, since under sensor drifting it is extremely difficult to distinguish mild slopes from flat ground. By this assumption, we cluster nearby SBPs with similar heights into buckets and store the mean height for each bucket. When a new SBP is detected, if an existing nearby cluster is similar in height, this SBP will join that cluster and update its mean height. $p = -k * d$ will be proposed to drag the root so that the SBP is closer to the updated mean height, where $k$ is a constant correction coefficient and $d$ is the difference between SBP height and cluster mean height.

Second, the terrain is assumed to be a Voronoi diagram with different vertical levels. That is, we assign each unvisited grid to the same height as its closest SBP-visited neighbor. To build Voronoi online, upon each new SBP visitation, we need to check if each unvisited grid is now closer to the new SBP than all existing SBPs. Instead of storing all distances, we only need to store the closest distance up to now, which we call (inverse) confidence map $\bm{C} \in \mathbb{R}^{L \times L}$. If the new SBP is closer, the height map $\bm{H}$ at the unvisited grid will be updated to the new SBP’s height (precisely, the corresponding SBP cluster’s height, from the first assumption).

Further implementation details can be found in Appendix C.

\section{Evaluations}




We organize our experiments in this section to demonstrate that:
\begin{itemize}
    \item Using the same held-out datasets of real or synthesized sparse IMU signals as benchmarks, our method improves over recent works quantitatively by a significant margin.
    \item Without training on any real non-flat-terrain IMU data, and without any ground-truth terrain supervision, our method can reconstruct motions on different terrains both in simulation and on real sensor data.
    \item Both real-time motion correction and terrain reconstruction using SBPs contribute to our ability to stably mitigating root drift during a variety of human activities.
    \item Modifying joint history with SBP correction mitigates drift between joints and qualitatively improves motion reconstruction in challenging motions.
    \item Our Transformer decoder model taking history as input facilities stable and consistent learning of the SBPs.
\end{itemize}



\subsection{Quantitative Evaluation on Flat Terrain}

We evaluate our results on existing IMU datasets that cover diverse types of motions and are paired with ground-truth full-body motions, for quantitative comparisons. 

\paragraph{Datasets}

Following previous works, evaluation datasets include DIP and TotalCapture of real IMU data. We match the exact training data and evaluation settings of baseline methods. This mainly serves to test the sim-to-real transfer of the models, which are all predominantly (${\sim}95\%$) trained on synthesized AMASS data (Appendix A). Additionally, we hold out a synthetic IMU dataset, DanceDB, to evaluate cross-motion-type generalization.

\begin{itemize}
    \item \textbf{DIPEval} \emph{(real heldout)}:  Data from two held-out subjects in the ten-subject DIP dataset.
    \item \textbf{TotalCapture} \emph{(real heldout)}: We held out real IMU measurements from the TotalCapture dataset~\cite{Trumble:BMVC:2017} for evaluation, but still use its ground truth and synthesized IMU readings as part of the AMASS training set, following the same practice as previous works.
    \item \textbf{DanceDB} \emph{(synthetic heldout)}: A large dataset of contemporary dances, therefore containing unique motion types to any other training dataset. Note that DanceDB is part of AMASS but we intentionally hold it out from our training data. Previous works likely did not include them in training either since they predate DanceDB's release in AMASS.
\end{itemize}


\paragraph{Metrics} 
We use the following metrics common for evaluating motion reconstruction quality. We randomly sample 600 consecutive frames (10s) from each motion in the evaluation datasets, to prevent very long motions biasing the statistics, and to avoid evaluation of root translation error in the beginning of motions where they mostly start from a stationary standing pose.

\begin{itemize}
    \item \textbf{Mean Joint Angle Error} \emph{(in degrees)}: Joint angle (represented in axis-angles) difference between reconstruction and ground-truth, averaged over all joints.
    \item \textbf{Mean Root-Relative Joint Position Error} \emph{(in centimeters)}: Global joint Cartesian position difference (Euclidean norm) between the reconstruction and ground-truth by aligning at the root, averaged over all joints.
    \item \textbf{Root Error 2s/5s/10s} \emph{(in meters)}: Root translation error measured in Euclidean norm during a continuous period of 2s/5s/10s. Note that existing works have no errors from the vertical axis as they are designed for flat-ground motions.
    \item \textbf{Mean Joint Position Jitter} \emph{(in $m/s^3$)}: Joint position jitter computed using the same formula as in TransPose, averaged over all joints.
    \item \textbf{Root Jitter} \emph{(in $m/s^3$)}: Root position jitter computed using the same formula as above.
\end{itemize}

\begin{table}[b]
\caption{\label{table:eval} 
 Comparison of model performance on evaluation datasets. Bold numbers indicate the best performing entries.
 }
\begin{tabular}{l}
\hspace*{-6pt}\includegraphics[width=0.48\textwidth]{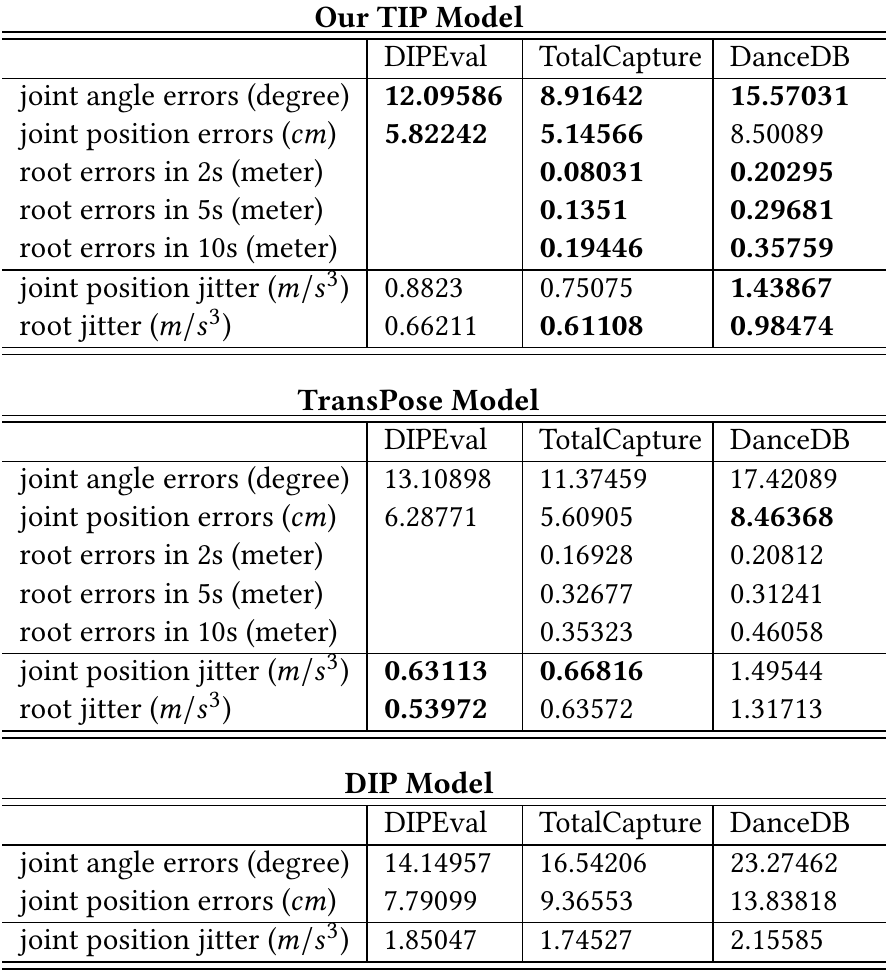}\\
\end{tabular}
\end{table}

\paragraph{Results}
We present quantitative metrics on the evaluation datasets between our model, TransPose, and DIP in Table~\ref{table:eval}. We used the best performing models published by the authors in this comparison. Overall, our system achieves better accuracy in almost all evaluations. Our system does not have an offline mode, and all reported metrics reported are from online inference (i.e. pretending existing data files as streaming in).

We see most significant improvements from previous works on root translation, on both TotalCapure and DanceDB datasets. Sec. \ref{sec:ablation} shows the effects of our key design choices on root drifting mitigation. Thanks to our prepossessing filtering (Appendix A), we can treat the DIP training split as ordinary ${\sim}5\%$ subset of the whole training data with the rest ${\sim}95\%$ being purely synthetic IMU. By using a simple one-stage training procedure, instead of the two-stage training-then-finetuning as in previous works, we reduce the risk of overfitting to activities in the DIP dataset during finetuning.

Comparing to results from concurrent PIP work \cite{PIP:CVPR:2022}, our method achieves a similar level of improvement over TransPose. On flat terrain activities, our work complements the PIP model by exploring drastically different techniques. We are not able to compare with this new work quantitatively as the code and model have not been released yet.

\subsection{Quality of Simultaneously Generated Terrains}

Since our reconstructed terrains are only one \textit{plausible} result among infinite possibilities, we perform the following two \textit{qualitative} experiments to evaluate their consistency with the human motion:

\begin{itemize}
    \item \textbf{Simulation}: we evaluate our terrain generation algorithm by running our system on terrain navigation motions included in the AMASS synthesized IMU training data. (Video 1m10s, note AMASS does not provide terrain ground-truths.)
    \item \textbf{Live Demos}: though our model has never seen any real IMU signals on terrains, we push its limit and showcase simultaneous motion reconstruction and terrain generation for walking on stairs (Video 1m56s), and climbing/jumping off chairs (Video 2m23s).
\end{itemize}

\begin{figure}[ht]
\centering
\includegraphics[width=0.9\linewidth]{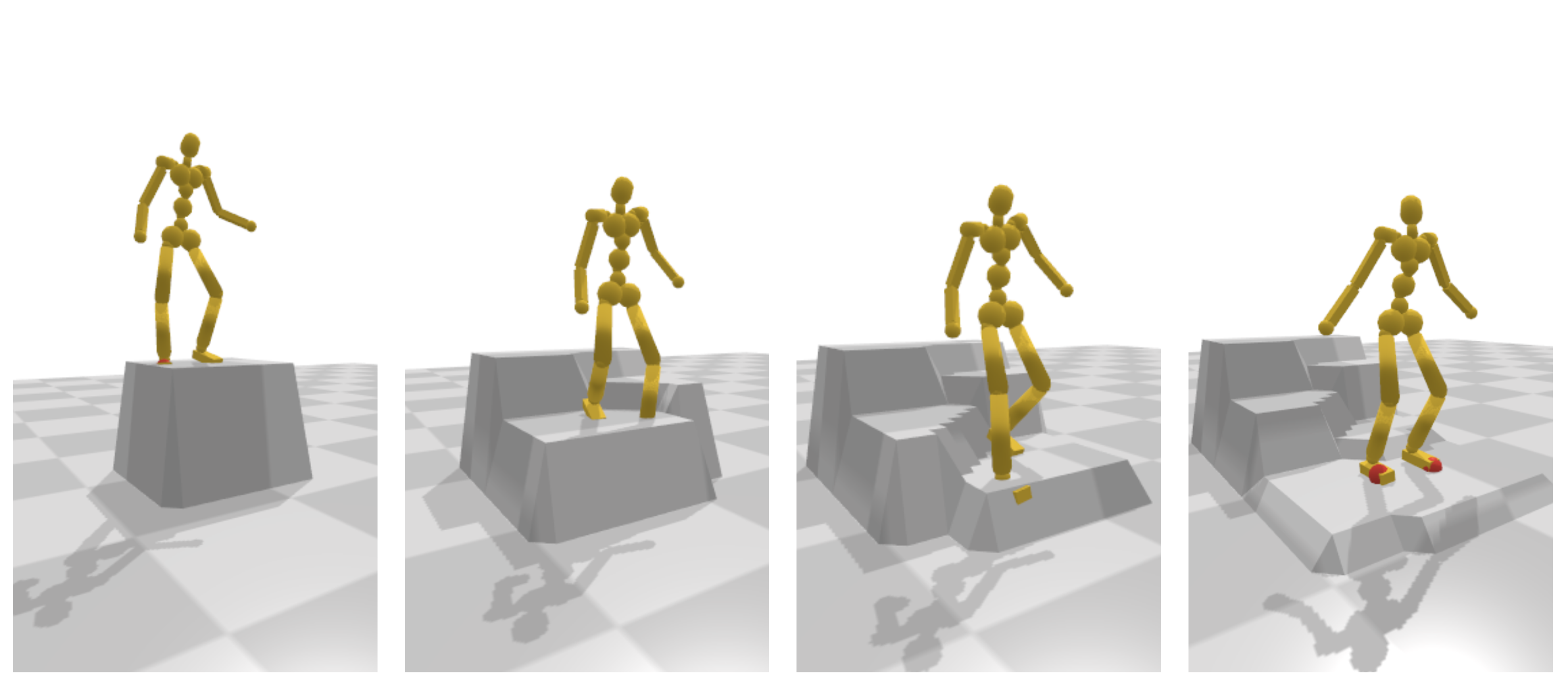}
\caption{Generated terrain. Many equally plausible solutions exist for same motion, the solution from our algorithm being one of them.}
\label{fig:generated-map}
\vspace{-0.4cm}
\end{figure}

\begin{table*}[ht]
    \caption{Comparison of ablation models on TotalCapture evaluation dataset.}
    \begin{tabular}{l|c|c|c|c|c}
         &  No history, w/ SBP & No SBPs & Predict SBPs wo/ usage & wo/ terrain corrections & TIP \\
\hline
\hline
    root errors in 2s	  & 0.15833	& 0.16987 & 0.13927	& {\bf 0.07817}	& {\bf 0.08031} \\
    root errors in 5s	  & 0.28575	& 0.32543 & 0.32972	& 0.15462	& {\bf 0.13509} \\
    root errors in 10s	  & 0.41114	& 0.5051 & 0.51706	& 0.24196	& {\bf 0.19445} \\

\hline
    \end{tabular}
    \label{table:ablation}
\vspace{-2mm}
\end{table*}
The supplementary video demonstrates plausible terrains both in live demos and on the AMASS dataset. \fig{fig:generated-map} presents one of the generated terrains from simulation. Although we do not have access to the ground truth terrain, our Voronoi-graph based algorithm generates plausible staircases that are consistent with the human motion. Body parts may still penetrate the terrain briefly while it is being generated (\fig{fig:generated-map}, second to the left), because the algorithm ignores each newly active SBP for a short period of time (Appendix C). Final terrain should see little penetration when playing back the motion.


\subsection{Ablation Studies on Root Drifting}
\label{sec:ablation}

Multiple key design choices improved our root translation accuracy, which is also crucial to the success of terrain generation. To gain a better understanding, we created the following ablation models:

\begin{itemize}
    \item \textbf{No history, with SBP}: a Transformer \textit{encoder} can be used as a sequence summarizing model, taking time windows of IMU signals and output only the current full-body pose $\bm{q}_t$, $\bm{v}_t$ and SBPs $\bm{c}_t$, with no consideration of its past predictions. We still used the SBPs to correct the root in this case.
    \item \textbf{No SBPs}: same network as Ours without learning SBPs.
    \item \textbf{Predict SBPs without Usage}: no test-time usage of the predicted SBPs. This is different from "No SBPs" since learning an unused auxiliary task may help the main task learning.
    \item \textbf{Without Terrain Corrections}: Do not use terrain algorithm for vertical drift correction. 
\end{itemize}

Table. \ref{table:ablation} summarizes the models' performances on the real-IMU TotalCapture dataset. Both "no SBPs" and "predicting SBPs without usage" rely on the model's raw prediction of $\bm{v}_t$, and perform the worst in this test, showing the importance of both learning the SBPs and enforcing them at run time. "No history, with SBP" also fails, interestingly because the SBPs predicted by a Transformer encoder becomes intermittent in this case and therefore less helpful during run time. This shows the importance of explicitly including output history as input. Intuitively, the onset history of SBPs are important information for next SBP onset prediction.

Finally, without terrain-based root drifting correction, in some testing motions, vertical drift may be large for reasons such as biased sensor error from calibrations. Such errors in vertical direction also negatively affects generated terrains if not corrected in time.

\subsection{Qualitative Comparisons of Run-time Joint IK Correction}
Introduced in Sec. \ref{sec:method-sbp}, we use IK to correct joint prediction histories with a pair of active SBPs. We showcase its usage in one of the most common joint drifting scenarios of long sitting \cite{TransPose:SIG:2021,PIP:CVPR:2022}, where the character may easily transition to standing over an extended period of time due to similar orientations and accelerations on legs and waist IMUs. With our SBP IK correction, we can stably reconstruct a two-minute sitting sequence, both with our own sensors (Video 0m42s) or using public real-IMU sequences (Video 3m10s) in DIP dataset, showing the value of correcting history buffer with IK to be consistent with predicted SBPs.



\subsection{Live Demo}
We test our system live with 6 Xsens IMU sensors. Our video visualizes live performance side-by-side with real-time reconstructions, with a slight latency caused by our pre-processing filter. Besides the aforementioned tasks, we cover a variety of motion tasks in our demos, both common ones such as locomotion and whole-body manipulation, and more challenging ones such as jumping from a high place, "swimming" on a stool, dancing, hand on floor movements, or swirl kicks. We tested our system on one male and one female subjects, and observed degraded performance on the female subject which we did not expect (Appendix F).

\section{Conclusion}
\label{sec:Conclusion}

This paper presents a new data-driven method for human motion reconstruction from six wearable IMUs, with simultaneous plausible terrain generation. By combining a conditional Transformer decoder model for consistent prediction, a hybrid drift stabilizer utilizing learned stationary information across human body, and an algorithm to simultaneously generate regularized terrain and correct noisy global motion estimation, new downstream applications can be made possible with this self-contained and economic setup of motion capture. For future work, our method could be largely improved by collecting real IMU datasets with motions on various terrains. Personalized finetuning or calibration may also improve reconstruction for each individual user.

\begin{acks}
To members of the Stanford Movement Lab, the Stanford Human Performance Lab, Meta Reality Labs Research, and to Josh Cooley, Yinghao Huang, Manuel Kaufmann, Ari Tamari,   Xia Wu, Di Xia, Xinyu Yi, Eris Zhang,  for helpful discussions and technical assistance. To anonymous reviewers whose feedback substantially helped refine this work. Yifeng Jiang is partially supported by the Wu Tsai Human Performance Alliance at Stanford University.
\end{acks}


\balance
\bibliographystyle{ACM-Reference-Format}
\bibliography{bibliography}

\clearpage

\newpage

\appendix

\section{Acceleration Readings and Sim-to-Real}
\label{sec:method-data}

\paragraph{Moving Average Filtering.} Since real IMU data paired with ground-truth full-body motions are small in size, following previous work \cite{DIP:SIGA:2018}, we place virtual IMU sensors on virtual characters driven by captured motions to synthesize IMU orientation and acceleration readings. Using the AMASS \cite{AMASS:ICCV:2019} motion dataset (a collection of smaller motion capture datasets), we create a large-scale synthetic IMU dataset for training our model.

\begin{figure}[h]
\centering
\includegraphics[width=\linewidth]{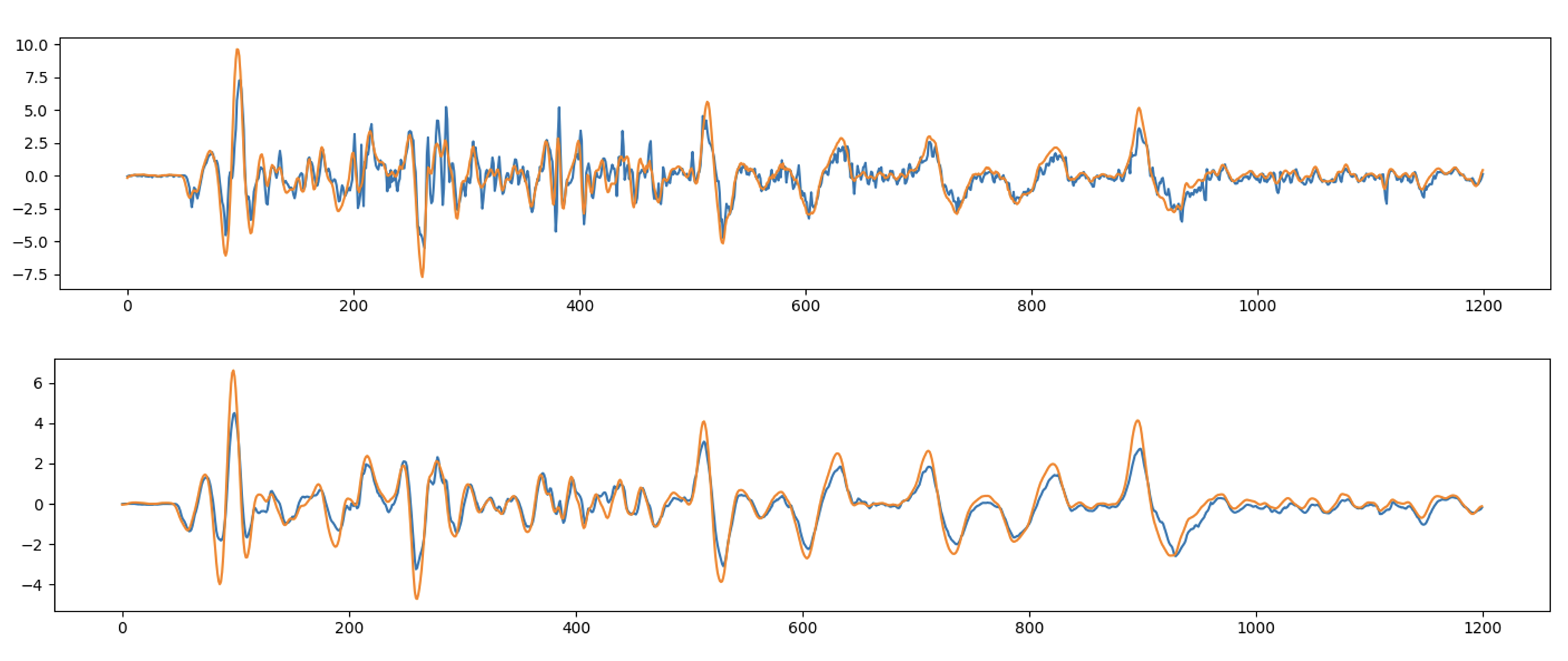}
\caption{Example of synthesized (orange) and real (blue) acceleration data, before (top) and after (bottom) moving average filtering.}
\label{fig:smooth}
\vspace{-5mm}
\end{figure}

However, synthetic and real IMU data exhibit vastly different noise profiles. Acceleration data in the real dataset are noisy, but not in the same way as the noise in the synthetic dataset, which is caused by double differentiation of mocap data (Figure \ref{fig:smooth} Top). On the other hand, orientation data are usually less noisy because they are processed by the in-sensor Kalman filter~\cite{kalman:1960:new}. Previous work~\cite{DIP:SIGA:2018, TransPose:SIG:2021} recognized this distribution mismatch problem and proposed to first train the model exclusively on the synthetic data and then finetune it on a smaller real dataset. This two-step solution leads to a more complex training procedure that requires careful tuning to avoid overfitting the real dataset.

We found that simply running an average filter on both synthetic and real acceleration data (with window length of 11 in our implementation) would bring the two data sources sufficiently close to each other (Fig. \ref{fig:smooth} Bottom). We then train the model only once on the combined dataset. Combining both data sources simplifies training from two stages to one stage, and avoids the risk of catastrophic forgetting during finetuning.

In practice, filtering causes latency during real-time inference, as computing moving average requires future IMU readings. We use 5 times steps (83ms) of future readings, the same requirement as \cite{TransPose:SIG:2021, DIP:SIGA:2018}, though they require future readings as part of model input while we merely use them for filtering.

\paragraph{Summing Up (Integrate) Past Accelerations.} Another issue we discover for non-flat terrain motions is that the sensor readings (both orientation and acceleration) during a stair step is much similar to a normal flat-ground step, especially in the case of real, noisy IMU data. Note that pelvis also accelerates up and down in a normal walking step resembling an inverted pendulum. However, if we sum up the raw IMU acceleration readings within a small window of recent history (e.g. past 0.5s), similar to "integrating" acceleration to delta velocities, we could observe a more different signal shape between stairs and normal steps. Empirically, we find adding this additional history sum features for each channel, increasing acceleration features from $\mathbb{R}^{18}$ to $\mathbb{R}^{36}$ (concatenating with filtered accelerations), improves stair recognition on real hardware.

\section{Model Details}
\label{sec:implementations-model}
We use the AMASS dataset to generate synthetic training data following the smoothing procedure in Appendix \ref{sec:method-data}. It consists of over a dozen different motion capture datasets performing a variety of activities. In addition, we include $8$ out of $10$ subjects' data from the DIP dataset. We use pyBullet \cite{coumans2016pybullet} for calculating forward kinematics during data synthesis, SBP label generation, root correction, and final visualizations. As the DIP real IMU data do not have root motion, we use a pre-trained model to label pseudo ground-truth SBPs for the DIP motions.


We use standard loss functions for the model outputs, i.e., mean-squared error for joint rotations, mean-squared error for $\bm{v}_t$ and Cartesian elements of $\bm{c}_t$, and binary cross-entropy for binary elements of $\bm{c}_t$, (i.e. $b_t$). Specifically, for joint rotations,
\begin{equation*}
    \mathcal{L}_J = \norm{\bm{q}_t - \bar{\bm{q}}_t}_2^2,
\end{equation*}
where $\bar{\bm{q}}_t$ is the ground-truth full-body joint rotations, represented as first two columns (6D) of each rotation matrix as noted previously in the main text. Similarly for root velocities,
\begin{equation*}
    \mathcal{L}_R = \norm{\bm{v}_t - \bar{\bm{v}}_t}_2^2,
\end{equation*}
and for the SBP predictions $\bm{c}_t$,
\begin{equation*}
    \mathcal{L}_C = \sum_{i=1}^5 \norm{\bm{r}_t^{(i)} - \bar{\bm{r}}_t^{(i)}}_2^2 + \sum_{i=1}^5 ( -\bar{b}_t^{(i)} \log b_t^{(i)} - (1 - \bar{b}_t^{(i)}) \log (1 - b_t^{(i)}) ).
\end{equation*}
Since our model during training time predicts a whole trajectory window, we experimented with a jerk loss penalizing deviation of neighboring frames, but it did not produce visible improvements. This might be due to the fact that during test time we still only use the last prediction at each step. Instead, we pass our output through an exponential moving average filter as post processing, at the expense of slight increase in joint accuracy errors (Appendix \ref{sec:discussions}).

Our model is trained in PyTorch~\cite{paszke:2019:pytorch} using the Adam optimizer~\cite{kingma:2014:adam}, with a batch size of 256 and a learning rate of 0.0001 multiplied with a cosine schedule \cite{loshchilov2016sgdr}. We perform training for 1000 epochs, which takes around 6 hours with a GeForce GTX 2080Ti GPU. Once trained, our model is small enough to run at $60$ fps on a 2080Ti machine, with bottleneck being the python wrapper of pyBullet. Our model uses max window size $M=39$. It contains a total number of $3,677,315$ parameters, comparing to $4,798,771$ in TransPose and $10,801,934$ in DIP.

Note that our model requires an initial full-body pose given in the first step of prediction. In practice this is always the case since the sensors need to be calibrated with a T pose before each use, as they are allowed to be slightly differently worn. See Appendix \ref{sec:calibration} for more details.

\section{Terrain Generation Details}
\label{sec:implementations-terrain}


Grid size is empirically set to 0.1m, and number of grids $L \times L$ is set so that terrain is large enough. We initialize the height map $\bm{H}$ with all zero values, where zero is set to the initial root height minus a constant height $w$. $w$ represents the lower bound of how low the subject could possibly reach in this capture. We assume no cluster means are below this value. If the user can provide a tighter $w$ (e.g. starting the capture on the lowest ground plane), we can generate a more visually pleasing terrain by producing no dents lower than the specified ground plane. During real-time demos, w is set to be tight, to indicate that we know the motion will not go lower than the starting ground plane.

If there were only two SBPs with different heights, Voronoi diagram will render a 1-step stair that is infinitely wide. For aesthetics, we limit the area each new SBP can influence to $1$m $\times 1$m, which could be nevertheless still wider than the real stairs in scene. While we can arbitrarily make this influence region narrower, we note that, without additional information, both are equally \textit{plausible}, and new SBPs can always crop the terrain narrower with more information streaming in (Video 1m40s).

For a newly active SBP, we ignore it for $t_0$ seconds before using it for the terrain algorithm, to allow it to settle in height. $t_0$ is $50$ frames or the the moment SBP becomes inactive, whichever comes earlier. The pelvis SBP is only used in terrain generation if it is $>0.2$m away from the feet, to avoid building terrains at the pelvis height when the subject is standing still. 

\section{Sensor Calibration}
\label{sec:calibration}


When testing on real hardware, as the raw sensor readings are in different coordinate frames from the frame of system input, calibration is needed to obtain the offset transforms between the coordinate frames beforehand. We adopt a slightly different IMU calibration procedure from previous works that is nevertheless still straightforward to explain. 

We start from defining a few coordinate frames. Let $G_n$ be the base (i.e. identity) frame of each of the $6$ sensors (for the Xsens sensors we used, identity orientation could mean different poses per sensor). Let $G_p$ be any fixed global frame the user specifies, whose $x$ axis indicates the specified front, $y$ axis corresponds to the left, and $z$ axis corresponds to the upwards. (Note this axis definition is different from DIP and TransPose models.) Let $S_t$ be the sensor frame, while $S_0$ defines the sensor frame during T-pose calibration. Let $B_t$ be the bone frame, while $B_0$ defines the bone frame during T-pose calibration. We omit the sensor indices $j$ (e.g. $G_n^{(j)}$, $S_t^{(j)}$) since calibration is agnostic to each sensor.

Using these notations, $\bm{R}_{G_n}^{S_t}$ represents the raw sensor orientation reading based from frame $G_n$, and $\bm{a}_{S_t}$ represents the raw acceleration reading which is always local in sensor frame. The system however expects both bone orientation and acceleration reading in $G_p$, i.e., $\bm{R}_{G_p}^{B_t}$ and $\bm{a}_{G_p}$. We have the following relations:

\begin{eqnarray*}
    & \bm{R}_{G_p}^{B_t} = \bm{R}_{G_p}^{G_n} \bm{R}_{G_n}^{S_t} \bm{R}_{S_t}^{B_t}, \\ 
    & \bm{a}_{G_p} = \bm{R}_{G_p}^{G_n} \bm{R}_{G_n}^{S_t} \bm{a}_{S_t} - \bar{\bm{a}}_{S_t},
\end{eqnarray*}
where we note that $\bar{\bm{a}}_{S_t}$ is the constant acceleration bias in global frame, usually just the gravitational acceleration. From these relations, it should be clear that the goal of calibration is simply to obtain $\bm{R}_{G_p}^{G_n}$ and $\bm{R}_{S_t}^{B_t}$ before each system run.

In the first calibration step, we place all sensors to align with the specified global frame so that $\bm{R}_{G_n}^{S} = \bm{R}_{G_n}^{G_p}$, and obtain $\bm{R}_{G_p}^{G_n}$ which is simply $\{\bm{R}_{G_n}^{S}\}^T$. Following \cite{TransPose:SIG:2021}, we keep all sensors still on ground for three seconds and take the average reading.

Next, to obtain $\bm{R}_{S_t}^{B_t}$, the user wears all six sensors and stand in a T pose, facing the same "front" as $G_p$. We assume that the sensor will stay static with respect to the bone throughout the entire system run, therefore $\bm{R}_{S_t}^{B_t} = \bm{R}_{S_0}^{B_0}$. Since the orientation of each bone at a standard T pose, $\bm{R}_{G_p}^{B_0}$, is known, we are able to obtain $\bm{R}_{S_0}^{B_0}$ from the T-pose raw sensor reading $\bm{R}_{G_n}^{S_0}$ using:

\begin{equation*}
    \bm{R}_{S_0}^{B_0} = \{\bm{R}_{G_n}^{S_0}\}^T \bm{R}_{G_n}^{G_p} \bm{R}_{G_p}^{B_0},
\end{equation*}
where same as the first step, T pose is maintained for three seconds and we use the average reading for $\bm{R}_{G_n}^{S_0}$.

\section{Additional Analysis}


We present results of two additional experiments in this section. First to showcase how much the performance our autoregressive model will degrade over time, we repeat the quantitative experiment of Table \ref{table:eval} but on random 3000-frame (50s) windows of each motion, instead of 600 frames (10s). Note that since many test motions are shorter than 50s, this experiment setting may unevenly bias statistics. For brevity, the DIP model is not included in this comparison:

\begin{table}[h]
\caption{\label{table:eval-50} 
 Comparison of model performance on evaluation motion segments of maximum length 50 seconds. Bold numbers indicate the best performing entries.
 }
 {\bf Our TIP Model}
 \begin{tabular}{l|p{35pt}|p{45pt}|p{40pt}}
\hline
\hline
   & DIPEval & TotalCapture & DanceDB \\
\hline
	joint angle errors (degree)    & {\bf 12.33555} & {\bf 9.46942} & {\bf 15.28491} \\
	joint position errors ($cm$)   & {\bf 5.86926} & {\bf 5.40289} & {\bf 8.23641} \\
	root errors in 2s (meter)   &         & {\bf 0.08545} & {\bf 0.09504} \\
    root errors in 5s (meter)       &         & {\bf 0.16679} & {\bf 0.20369} \\
	root errors in 10s (meter)      &         & {\bf 0.20338} & {\bf 0.38935} \\ 
\hline
	joint position jitter ($m/s^3$)  & 0.84848 & 0.80672 & {\bf 1.39043} \\
	root jitter ($m/s^3$)           & 0.64593  & {\bf 0.64609}  & {\bf 0.95740} \\
\hline
\hline
\end{tabular}

 \vspace{0.3cm}

 {\bf TransPose Model}
 \begin{tabular}{l|p{35pt}|p{45pt}|p{40pt}}
\hline
\hline
   & DIPEval & TotalCapture & DanceDB \\
\hline
	joint angle errors (degree)    & 12.78403 & 11.56577 & 17.22182 \\
	joint position errors ($cm$)   & 6.16507 & 5.76287 &  8.35314 \\
	root errors in 2s (meter)   &         & 0.18543 & 0.14899 \\
    root errors in 5s (meter)       &         & 0.32042 & 0.28216 \\
	root errors in 10s (meter)      &         & 0.32111 & 0.45332 \\ 
\hline
	joint position jitter ($m/s^3$)  & {\bf 0.57619} & 	{\bf 0.76578} & 1.44662 \\
	root jitter ($m/s^3$)           & {\bf 0.49804}  & 0.70235  & 1.29385 \\
\hline
\hline
\end{tabular}

\end{table}

Reading the numbers from Table \ref{table:eval-50}, degradation of model performance is minimal on longer motions, and the statistics trends between our model and TransPose remain unchanged. As a side note, the DanceDB dataset contains more short motions, making the sampling a random 50s segment more likely to cover the beginnings of motions. We therefore see both TIP and TransPose have improved \textit{root errors in 2s} since the motions usually start from standing and are less dynamic in the first two seconds.

Second, to showcase the benefit of acceleration preprocessing, we perform an ablation study where we remove the average filtering and summation operations from our TIP system, both during training and test time.

\begin{table}[h]
\caption{\label{table:eval-acc} 
 Ablation of model performance on evaluation motion segments of 10s. Bold numbers indicate the best performing entries.
 }
 {\bf Our TIP Model}
 \begin{tabular}{l|p{35pt}|p{45pt}|p{40pt}}
\hline
\hline
   & DIPEval & TotalCapture & DanceDB \\
\hline
	joint angle errors (degree)    & {\bf 12.09586} & {\bf 8.91642} & {\bf 15.57031} \\
	joint position errors ($cm$)   & {\bf 5.82242} & {\bf 5.14566} & {\bf 8.50089} \\
	root errors in 2s (meter)   &         & {\bf 0.08031} & 0.20295 \\
    root errors in 5s (meter)       &         & {\bf 0.1351} & 0.29681 \\
	root errors in 10s (meter)      &         & {\bf 0.19446} & 0.35759 \\ 
\hline
	joint position jitter ($m/s^3$)  & 0.8823 & {\bf 0.75075} & {\bf 1.43867} \\
	root jitter ($m/s^3$)           & 0.66211  & {\bf 0.61108}  & {\bf 0.98474} \\
\hline
\hline
\end{tabular}

 \vspace{0.3cm}
 
 {\bf Our TIP Model, w/o Acceleration Preprocessing}
 \begin{tabular}{l|p{35pt}|p{45pt}|p{40pt}}
\hline
\hline
   & DIPEval & TotalCapture & DanceDB \\
\hline
	joint angle errors (degree)    & 13.02724 & 9.18290 & 15.67625 \\
	joint position errors ($cm$)   & 6.35219 & 5.29268 & 8.58611 \\
	root errors in 2s (meter)   &         & 0.09096 & {\bf 0.16825} \\
    root errors in 5s (meter)       &         & 0.18015 & {\bf 0.25855} \\
	root errors in 10s (meter)      &         & 0.20726 & {\bf 0.34444} \\ 
\hline
	joint position jitter ($m/s^3$)  & {\bf 0.85845} & 0.79335 & 1.44460 \\
	root jitter ($m/s^3$)           & {\bf 0.64661}  & 0.63560  & 1.00417 \\
\hline
\hline
\end{tabular}

\end{table}

From Table \ref{table:eval-acc}, We see a visible improvement from preprocessing the raw acceleration readings on real-IMU datasets (DIPEval \& TotalCapture). As expected, preprocessing is unimportant for synthesized IMU data (DanceDB).

\begin{figure}[h]
\centering
\includegraphics[width=\linewidth]{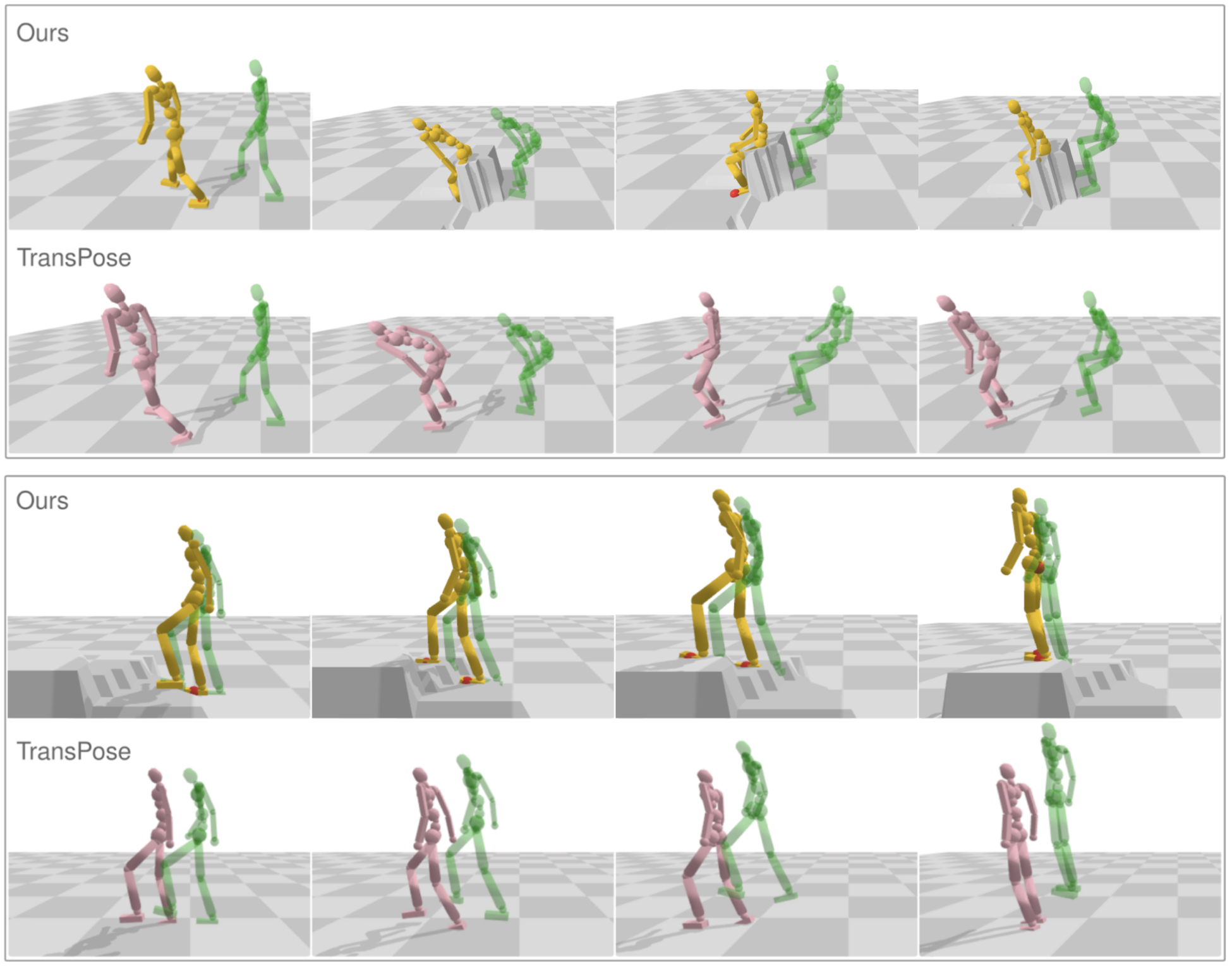}
\caption{Motion reconstruction for sitting on a chair (from real IMU data, top) and climbing steps (from synthesized IMU data, bottom). Our character is shown in yellow, TransPose in purple and Ground-Truth motion is shown in green. The red spheres are predicted SBPs. Playing back motion on reconstructed terrains.}
\label{fig:res_motions}
\end{figure}

Qualitative comparisons between our method and TransPose are presented using the following two representative motions (\fig{fig:res_motions}, Video 3m33s). Our TIP model can generate a more stable sitting posture by making better use of its own past predictions and utilizing run-time IK correction (\fig{fig:res_motions} Top). \fig{fig:res_motions} Bottom shows that our algorithm is terrain agnostic while TransPose assumes a flat ground and uses this assumption to correct the algorithm's vertical root prediction. 

\section{Discussions}
\label{sec:discussions}
Though we have shown clear improvement on existing challenges of temporal consistency due to ambiguity, dynamic motion coverage, and terrain coverage, our system still has a few drawbacks for future work. First, it tends to underestimate the terrain height rather than overestimate (e.g. Video 3m52s) - collecting more annotated real IMU data on various terrain types, and increasing training samples with uneven terrains through data upsampling, could both help improve terrain reconstruction. Second, terrain height estimations remain challenging since they solely depend on motion prediction, and are susceptible to sensor noises. For example, locomotion on a slightly bumpy ground versus on a flat ground is theoretically near-ambiguous given IMU's noise level (Video 4m6s). Third, our motion reconstruction quality on real hardware can degrade on motion types that are rare in training, thus affecting the quality of generated terrains (Video 4m21s). Finally, though our work does not claim contribution over the jitter level of reconstructed motions, the smoothing filter during post-processing is far from ideal and hurts our motion accuracy by effectively increasing latency.

Another very visible problem we observe is the model’s bias to body types. Our synthesized data were generated from virtual characters with random heights sampled from 1.6m to 1.8m. We observed that the algorithm generalizes better to taller users than shorter ones. We hypothesize that this phenomenon is due to the magnitude of acceleration, as the model might be more easily confused by smaller signals from a shorter user. Similarly, existing real IMU datasets might have a bias in human shapes. Some personalized training and finetuning of the model may eventually be necessary for reconstructing more accurate and detailed motion for each individual user.

The terrain generated from our algorithm is "plausible" in the sense that it cannot distinguish, sorely from IMU readings, if the foot is resting on a terrain or simply staying stationary in air (Video 4m58s). An algorithm that takes the distribution of commonly seen environments into consideration could guide our system to generate more likely terrains in such ambiguous cases.

\end{document}